%% file: main.tex
\documentclass[10pt,twocolumn,letterpaper]{article}

\usepackage[pagenumbers]{cvpr} 

\input{preamble}

\definecolor{cvprblue}{rgb}{0.21,0.49,0.74}
\usepackage[pagebackref,breaklinks,colorlinks,allcolors=cvprblue]{hyperref}

\def\paperID{3503} 
\def\confName{CVPR}
\def\confYear{2026}

\title{Learning to Generate via Understanding: Understanding-Driven Intrinsic Rewarding for Unified Multimodal Models}

\author{Jiadong Pan$^{1,2}$~~~
Liang Li$^{1}$\footnotemark[1]~~~
Yuxin Peng$^{3}$~~~
Yu-Ming Tang$^{4}$~~~
Shuohuan Wang$^{5}$\\~~~
Yu Sun$^{5}$~~~
Hua Wu$^{5}$~~~
Qingming Huang$^{1,2}$~~~
Haifeng Wang$^{5}$
\\
	$^1$Institute of Computing Technology, Chinese Academy of Sciences~~~$^3$Peking University\\
    $^2$University of Chinese Academy of Sciences~~~$^4$Sun Yat-sen University~~~$^5$Baidu Inc.\\
    }

\begin{document}
\maketitle

\renewcommand{\thefootnote}{\fnsymbol{footnote}}
\footnotetext[1]{Corresponding author.}
\renewcommand{\thefootnote}

\input{sec/0_abstract}

\input{sec/1_intro}
\input{sec/2_relate_work}

\input{sec/3_preliminary}
\input{sec/4_methods}
\input{sec/5_experiments}
\input{sec/6_conclusion}

{
    \small
    \bibliographystyle{ieeenat_fullname}
    \bibliography{main}
}


\end{document}

%% file: preamble.tex
\usepackage{booktabs}
\usepackage{multirow}
\usepackage[dvipsnames]{xcolor}
\usepackage{colortbl}
\usepackage[normalem]{ulem}
\useunder{\uline}{\ul}{}

%% file: sec/0_abstract.tex
\begin{abstract}

Recently, unified multimodal models (UMMs) have made remarkable progress in integrating visual understanding and generation, demonstrating strong potential for complex text-to-image (T2I) tasks. 
Despite their theoretical promise, a persistent capability gap exists: UMMs typically exhibit superior visual understanding but comparatively weaker generative capabilities.
This discrepancy arises largely from the intrinsic decoupling between the understanding and generation processes.
While a UMM can accurately interpret fine-grained visual details, it often struggles to produce semantically coherent images from complex textual prompts.
To address this challenge, we explore UMMs' internal understanding capability to enhance generation quality.
We propose a token-level intrinsic text-image alignment reward mechanism, \textbf{GvU}, enabling the UMM to act simultaneously as teacher and student: it evaluates its own outputs using the understanding branch to guide the generations accordingly. Building upon this, we design a self-supervised reinforcement learning framework, allowing UMMs to iteratively improve their generation quality through understanding-based intrinsic reward signals—without reliance on external supervision.
Experimental results show that our method substantially boosts UMMs' generation, which in turn strengthens their fine-grained visual understanding, narrowing the capability gap between UMMs' visual understanding and generation. The project page is \url{https://matrix0721.github.io/gvu.github.io/}.

\end{abstract}

%% file: sec/1_intro.tex
\vspace{-6pt}
\section{Introduction}
\label{sec:intro}

The recent emergence of unified multimodal models (UMMs) represents a promising direction in artificial intelligence, aiming to consolidate visual understanding and generation into a unified task specification~\cite{team2024chameleon,zhou2024transfusion,chen2025blip3o,xie2025showo2,yan2025can,wu2025omnigen2,deng2025bagel,wu2025janus,chen2025januspro,wang2024emu3,pan2025transfer,ge2024seed,tong2025metamorph,dong2023dreamllm,shi2024lmfusion}. By sharing a unified backbone for both visual understanding and generation, UMMs offer a promising pathway toward complex instruction-following text-to-image (T2I) tasks. Superior to traditional T2I models~\cite{rombach2022high, ramesh2021zero,esser2024scaling,flux2024}, UMMs' advanced visual comprehension provides the potential to generate images that accurately reflect intricate spatial relationships, quantitative reasoning, and fine-grained attribute bindings~\cite{ghosh2023geneval,hu2024dpg}.

\begin{figure}[t]
    \centering
    \includegraphics[width=1.0\linewidth]{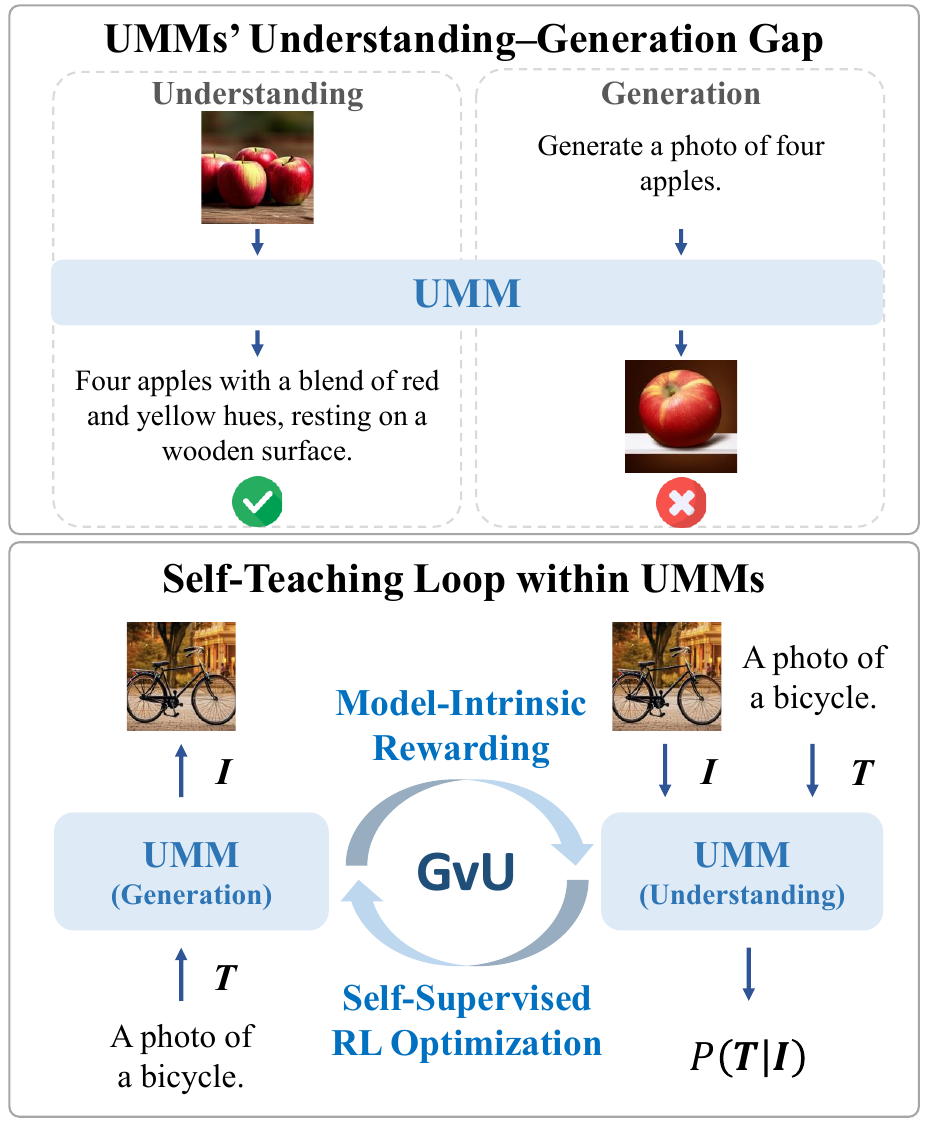}
    \vspace{-18pt}
    \caption{(a) Top: UMMs exhibit an understanding–generation gap: 
    they recognize visual details but fail to reflect them in generated images.
    (b) Bottom: our self-teaching mechanism uses intrinsic rewards from the understanding branch to guide generation, improving text–image alignment without external supervision.  
    }
    
    \label{fig_intro}
\vspace{-12pt}
\end{figure}

Despite the ambitious goals of UMMs, a severe gap between their theoretical potential and empirical performance reveals an imbalance in visual understanding and generation.
Current UMMs typically excel at visual understanding but lag in generation~\cite{chen2025januspro,pan2025transfer,wang2024emu3,wu2024vila}, partly because training pipelines prioritize understanding, while generative capabilities receive less supervision.
Joint training of UMMs for both understanding and generation further complicates this issue. Recent studies~\cite{team2024chameleon,fan2025unified,wu2025janus,pan2025transfer} show that optimizing two tasks together causes negative transfer, where gains in one task hinder the other. 
This antagonism challenges the core goal of UMMs—mutual enhancement across multimodal tasks—making the promotion of cross-task synergy between understanding and generation crucial for effective UMM training.

This challenge stems from a fundamental division in UMMs: visual understanding and generation are usually trained independently, with sparse integration of information between them. As a result, UMMs exhibit a performance asymmetry: visual understanding modules surpass visual generation counterparts. 
Conceptually, the two tasks—visual understanding (mapping images to text) and visual generation (mapping text to images)—are dualistic. 
This raises the possibility that UMMs’ visual understanding can be harnessed to enhance generative capability.
In T2I tasks, the text input contains the full semantic blueprint of the desired image. The understanding branch can assess how well a generated image aligns with the conditioning text, providing internal feedback without external supervision.
This observation suggests a self-evolving paradigm where the UMM teaches itself—the understanding branch guiding as the 'teacher' and the generation branch improving as the 'student'.
By embedding the self-teaching dynamic, UMMs can transfer knowledge from understanding to generation, transforming a source of antagonism into a catalyst for holistic multimodal synergy that progressively enhances generative quality.

Building upon the self-teaching paradigm, we design a self-supervised reinforcement learning (RL) framework to bridge the gap between generation and understanding in UMMs, enabling self-improvement on T2I tasks. 
Central to this framework, we propose a token-level text-image alignment intrinsic rewarding mechanism, \textbf{Generate via Understanding} (\textbf{GvU}), which uniquely leverages UMMs' own visual comprehension as a source of internal guidance for visual generation.
Specifically, GvU extracts token-wise text-image alignment logits from the understanding branch and uses them as fine-grained rewards to guide the generation branch. 
Unlike traditional image-level rewards~\cite{hessel2021clipscore,lin2024evaluating,zhang2023blind,xu2023imagereward}, the dense and semantically aligned signals allow UMMs to refine generation at a much finer granularity, capturing subtle semantic details. 
By iteratively enforcing token-level semantic consistency, the understanding branch of UMMs acts as an internal evaluator, guiding the generation branch toward greater semantic accuracy and enabling self-improvement via internal supervision.

Extensive experiments are conducted across multiple benchmarks.
Notably, on challenging text–image alignment tasks, such as GenEval++~\cite{ye2025echo}, our approach achieves 43.3\% improvement over the base model. 
During RL, the intrinsic reward consistently increases, indicating stronger semantic alignment. Furthermore, we observe improvements in fine-grained visual understanding, indicating that enhanced generation in turn promotes comprehension.


Our key contributions can be summarized as follows:
\begin{itemize}
    \item We propose a token-level text-image alignment intrinsic reward in UMMs that enables internal evaluation of fine-grained text–image semantic correspondences.
    \item We design a self-supervised reinforcement learning framework for UMMs that intrinsically leverages visual understanding to guide generation, effectively bridging the gap between the two capabilities.
    \item Our method achieves superior performance in visual generation tasks, and enhanced generation in turn improves UMMs’ fine-grained visual understanding, demonstrating self-supervised RL’s potential to advance both.
\end{itemize}

%% file: sec/2_relate_work.tex
\section{Related Work}

\begin{figure*}[!htbp]
    \centering
    \includegraphics[width=1.0\linewidth]{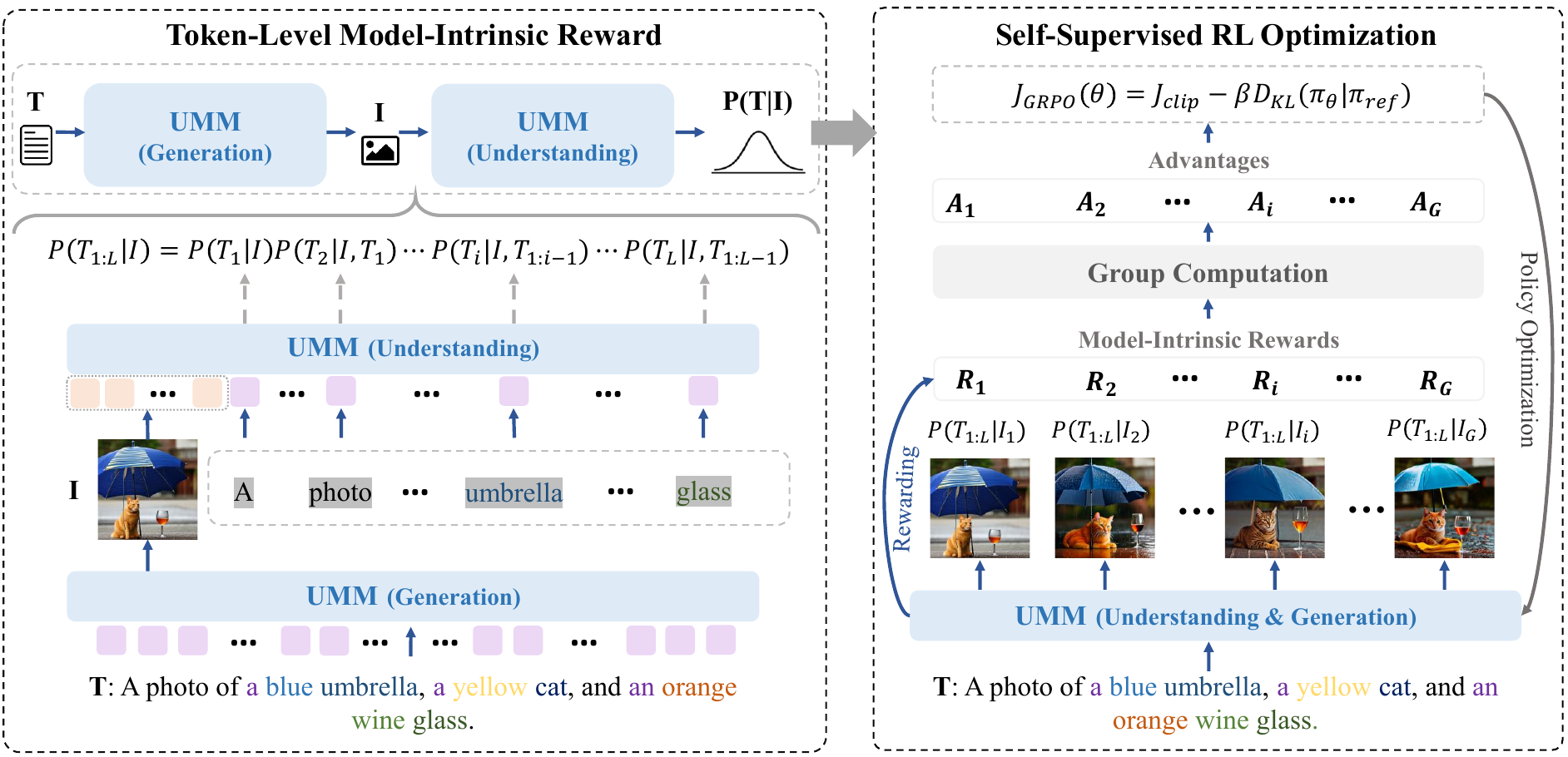}
    \vspace{-18pt}
    \caption{Overview of the GvU implementation. It comprises two key components: the \textit{token-level model-intrinsic reward} that provides fine-grained text–image alignment signals, and the \textit{self-supervised reinforcement learning (RL) process} that uses these signals to enhance UMMs' generative ability progressively, enabling continuous self-improvement without external supervision.}
    \label{fig:method}
\vspace{-12pt}
\end{figure*}

\subsection{Unified Multimodal Models}

Unified Multimodal Models (UMMs) have emerged as a prominent research paradigm in multimodal learning, enabling both visual understanding and generation through shared parameters and architecture.
The development of UMMs can be broadly classified into three architectural paradigms: (1) Autoregressive (AR) architectures (e.g., Chameleon~\cite{team2024chameleon}, Emu3~\cite{wang2024emu3}, Janus~\cite{wu2025janus}), which model visual and textual tokens sequentially; (2) Diffusion-based architectures (e.g., Bagel~\cite{deng2025bagel}, Show-o~\cite{xie2024show}), which synthesize images through iterative denoising processes; and (3) Hybrid architectures (e.g., Blip-3o~\cite{chen2025blip3o}, X-Omni~\cite{geng2025xomni}), which integrate AR modeling with diffusion-based decoding.
Compared with purely AR or diffusion-based designs, hybrid architectures assign image reconstruction to a diffusion head, decoupling semantic learning from low-level reconstruction and enabling more efficient integration of understanding and generation.
However, only a few studies~\cite{yan2025uae, xie2025reca} have begun to explore such hybrid enhancement.
Our method builds on hybrid framework to align understanding and generation, enhancing the model’s performance on complex T2I tasks.


\subsection{Reinforcement Learning of UMMs}

Reinforcement learning (RL) has been successfully applied to the training of LLMs~\cite{guo2025deepseek,rafailov2023direct,yang2025qwen3,yuan2024self} and is increasingly being extended to UMMs~\cite{fang2025got,duan2025gotr1,guo2025can}, where reward signals like text–image alignment metrics~\cite{hessel2021clipscore,lin2024evaluating} and human preference ratings~\cite{xu2023imagereward,wu2023hpsv2} guide image generation.
However, these reward functions are prone to reward hacking and lack fine-grained evaluation granularity, limiting their effectiveness in capturing subtle semantic details.
Despite the recent progress of UMMs, self-supervised learning for UMMs remains relatively underexplored. RecA~\cite{xie2025reca} proposes a post-training approach centered on reconstruction loss.
Some works~\cite{jin2025srum,hong2025reinforcing,han2025turning,mao2025unirl} incorporate rewarding mechanisms for UMMs to improve generation or understanding.
Nonetheless, fine-grained evaluation and a more systematic examination of the relationship between generative and understanding capabilities are still underexplored.
In contrast, our approach harnesses the model’s inherent visual understanding capability to improve generation, designing a self-supervised reinforcement learning framework that explicitly aligns generation with comprehension.


%% file: sec/4_methods.tex
\section{Methods}


To enhance UMMs’ generation via internal understanding, we construct a training pipeline integrating data generation, reward computation, and self-supervised reinforcement learning. 
First, we develop a self-generation data pipeline that forms a closed training loop within the model without relying on external image resources (Sec.~\ref{sec_method_data_pipeline}).
Next, we introduce token-level model-intrinsic reward, which provides intrinsic probability of image–prompt pairs to capture alignment between visual content and text (Sec.~\ref{sec_method_reward}).
Finally, we implement a self-supervised reinforcement learning procedure based on GRPO algorithm~\cite{shao2024deepseekmath}, enabling the UMM to progressively refine text–image alignment and improve instruction-following in complex T2I scenarios (Sec.~\ref{sec_method_rl}).


\subsection{Self-Generation Pipeline in UMMs} \label{sec_method_data_pipeline}

We adopt the UMM based on an autoregressive backbone with a diffusion head and develop a self-image generation and evaluation pipeline. Concretely, given only a dataset $\mathcal{D}_T$ of text prompts, the model takes these prompts as input on the generation branch to produce image tokens, which are then decoded into images through the diffusion head. In parallel, on the understanding branch, the model processes the generated images together with the original prompts to yield token-level intrinsic rewards. This entire process eliminates the need for external models or additional image datasets, forming a closed-loop generation–evaluation cycle that improves data efficiency and computational autonomy.

Specifically, during image generation, given text prompt tokens 
$T = T_{1:L} \sim \mathcal{D}_T$, 
the UMM $\theta$ first autoregressively models the causal distribution over text tokens:
\begin{equation}
    p_\theta(T_{1:i}) = \prod_{j=1}^i p_\theta(T_j \mid T_{j<i}),\quad i=1,2,\cdots,L.
\end{equation}
Conditioned on the complete text sequence $T_{1:L}$, the model then generates image tokens causally:
\begin{equation}
    p_\theta(I_{1:L_I} \mid T) = \prod_{j=1}^{L_I} p_\theta(I_j \mid T_{1:L}, I_{j<i}).
\end{equation}
The resulting image tokens $I_{1:L_I}$ are subsequently decoded into a pixel-space image through the diffusion head.
During image understanding, given system instruction tokens $P_{\text{sys}}$ 
and generated image tokens $I = I_{1:L_I}$, 
the UMM autoregressively models the text output distribution:
\begin{equation}
    p_\theta(\hat{T}_{1:i}) = \prod_{j=1}^i p_\theta(\hat{T}_j \mid P_{sys}, I_{1:L_I}, \hat{T}_{j<i}).
\end{equation}
Instead of sampling new text, the original text prompts are used as ground truth to compute their 
\textit{model-intrinsic probability} of the provided tokens, reflecting the model’s internal alignment between visual and textual modalities.

\subsection{Token-level Model-Intrinsic Reward}
\label{sec_method_reward}


To quantitatively evaluate the alignment between generated images and text prompts, we propose a token-level probability reward that exploits intrinsic generation probabilities of UMMs. The reward computes the likelihood of the ground-truth prompt tokens conditioned on generated images to measure image–text alignment. 


Given a text prompt $T=T_{1:L}$ and an image $I$ generated by the UMM $\theta$, the image is first converted into a discrete representation compatible with the autoregressive architecture and concatenated with special instruction tokens to form the model input sequence: $\textbf{X}_0=[I]$. 
The model then computes the token-level probabilities of the sequence $T_{1:L}$ autoregressively. Specifically, the probability of the j-th token $T_j$ is given by:
\begin{equation}
    p_\theta(T_j|\textbf{X}_{j-1})= \text{Softmax}(\text{Logits}_\theta(\textbf{X}_{j-1}) [T_j],
\end{equation}
where $\text{Logits}_\theta(\cdot)$ produces the logits for all possible next tokens given the current input sequence $\textbf{X}_{j-1}$ and the index $[T_j]$ selects the logit corresponding to the ground-truth token $T_j$. After computing the probability, the input sequence is updated by appending the current token $T_j$: $\textbf{X}_j= [\textbf{X}_{j-1};T_j]$.
The overall probability is computed as the geometric mean of all token probabilities to mitigate length bias:
\begin{equation}
    P(T_{1:L}|I)= (\prod_{j=1}^L p_\theta(T_j|\textbf{X}_{j-1}))^{1/L}.
\end{equation}
$P(T_{1:L}|I)$ represents the model’s intrinsic probability of the image–text pair $\{T,I\}$, reflecting their visual–textual alignment. This probability serves as a fine-grained intrinsic reward signal for text–image alignment, enabling self-supervised reinforcement learning in UMMs.

\begin{table*}[]
\caption{Evaluation of text-to-image generation capability on GenEval~\cite{ghosh2023geneval} benchmark. "\dag" refers to the method using the LLM prompt rewriter. \textbf{Bold} marks the best performance, and {\ul underlined} marks the second best. }
\vspace{-9pt}
\centering
\resizebox{\textwidth}{!}{
\begin{tabular}{@{}c|c|cccccc|c@{}}
\toprule
Model Type & Method & Single Obj. $\uparrow$ & Two Obj. $\uparrow$ & Counting $\uparrow$ & Colors $\uparrow$ & Position $\uparrow$ & Attr. Binding $\uparrow$ & Overall $\uparrow$ \\ \midrule
 & LDM~\cite{rombach2022high} & 0.92 & 0.29 & 0.23 & 0.70 & 0.02 & 0.05 & \cellcolor[HTML]{EFEFEF}0.37 \\
 & SD v1.5~\cite{rombach2022high} & 0.97 & 0.38 & 0.35 & 0.76 & 0.04 & 0.06 & \cellcolor[HTML]{EFEFEF}0.43 \\
 & SD v2.1~\cite{rombach2022high} & 0.98 & 0.51 & 0.44 & 0.85 & 0.07 & 0.17 & \cellcolor[HTML]{EFEFEF}0.50 \\
 & SD XL~\cite{podell2023sdxl} & 0.98 & 0.74 & 0.39 & 0.85 & 0.15 & 0.23 & \cellcolor[HTML]{EFEFEF}0.55 \\
 & IF-XL~\cite{deepfloyd2023if} & 0.97 & 0.74 & 0.66 & 0.81 & 0.13 & 0.35 & \cellcolor[HTML]{EFEFEF}0.61 \\
 & DALL-E~\cite{ramesh2022hierarchical} & 0.96 & 0.87 & 0.47 & 0.83 & 0.43 & 0.45 & \cellcolor[HTML]{EFEFEF}0.67 \\
 & SD3-medium~\cite{esser2024scaling} & 0.99 & 0.94 & 0.72 & 0.89 & 0.33 & 0.60 & \cellcolor[HTML]{EFEFEF}0.74 \\
 & SD3.5-large~\cite{esser2024scaling} & 0.98 & 0.89 & 0.73 & 0.83 & 0.34 & 0.47 & \cellcolor[HTML]{EFEFEF}0.71 \\
\multirow{-9}{*}{Only Generation} & FLUX.1-dev~\cite{flux2024} & 0.99 & 0.81 & 0.79 & 0.74 & 0.20 & 0.47 & \cellcolor[HTML]{EFEFEF}0.67 \\ \midrule
 & Show-o~\cite{xie2024show} & 0.95 & 0.52 & 0.49 & 0.82 & 0.11 & 0.28 & \cellcolor[HTML]{EFEFEF}0.53 \\
 & Janus~\cite{wu2025janus} & 0.97 & 0.68 & 0.30 & 0.84 & 0.46 & 0.42 & \cellcolor[HTML]{EFEFEF}0.61 \\
 & Janus-Pro~\cite{chen2025januspro} & 0.99 & 0.89 & 0.59 & 0.90 & 0.79 & 0.66 & \cellcolor[HTML]{EFEFEF}0.80 \\
 & Emu3~\cite{wang2024emu3} & 0.98 & 0.71 & 0.34 & 0.81 & 0.17 & 0.21 & \cellcolor[HTML]{EFEFEF}0.54 \\
 & UniWorld-V1~\cite{lin2025uniworld} & 0.99 & 0.93 & 0.79 & 0.89 & 0.49 & 0.70 & \cellcolor[HTML]{EFEFEF}0.80 \\
 & BLIP3-o 8B~\cite{chen2025blip3o} & - & - & - & - & - & - & \cellcolor[HTML]{EFEFEF}\textbf{0.84} \\
 & OmniGen2~\cite{wu2025omnigen2} & 1.00 & 0.95 & 0.64 & 0.88 & 0.55 & 0.76 & \cellcolor[HTML]{EFEFEF}0.80 \\
 & BAGEL~\cite{deng2025bagel} & 0.99 & 0.94 & 0.80 & 0.87 & 0.64 & 0.63 & \cellcolor[HTML]{EFEFEF}{\ul 0.81} \\
 & XOmni~\cite{geng2025xomni} & 1.00 & 0.94 & 0.60 & 0.85 & 0.40 & 0.26 & \cellcolor[HTML]{EFEFEF}0.68 \\
 & XOmni\dag~\cite{geng2025xomni} & 0.99 & 0.95 & 0.69 & 0.90 & 0.51 & 0.37 & \cellcolor[HTML]{EFEFEF}0.74 \\
 & GPT-4o~\cite{hurst2024gpt4o} & 0.99 & 0.92 & 0.85 & 0.92 & 0.75 & 0.61 & \cellcolor[HTML]{EFEFEF}\textbf{0.84} \\
 & \cellcolor[HTML]{EFEFEF}GvU (Ours) & \cellcolor[HTML]{EFEFEF}1.00 & \cellcolor[HTML]{EFEFEF}0.96 & \cellcolor[HTML]{EFEFEF}0.74 & \cellcolor[HTML]{EFEFEF}0.92 & \cellcolor[HTML]{EFEFEF}0.61 & \cellcolor[HTML]{EFEFEF}0.58 & \cellcolor[HTML]{EFEFEF}{\ul 0.81} \\
\multirow{-13}{*}{\begin{tabular}[c]{@{}c@{}}Generation\&\\ Understanding\end{tabular}} & \cellcolor[HTML]{EFEFEF}GvU (Ours)\dag & \cellcolor[HTML]{EFEFEF}1.00 & \cellcolor[HTML]{EFEFEF}0.97 & \cellcolor[HTML]{EFEFEF}0.80 & \cellcolor[HTML]{EFEFEF}0.93 & \cellcolor[HTML]{EFEFEF}0.68 & \cellcolor[HTML]{EFEFEF}0.65 & \cellcolor[HTML]{EFEFEF}\textbf{0.84} \\ \bottomrule
\end{tabular}

}
\label{table:geneval}
\vspace{-10pt}
\end{table*}

\begin{figure*}[]
    \centering
    \includegraphics[width=1.0\linewidth]{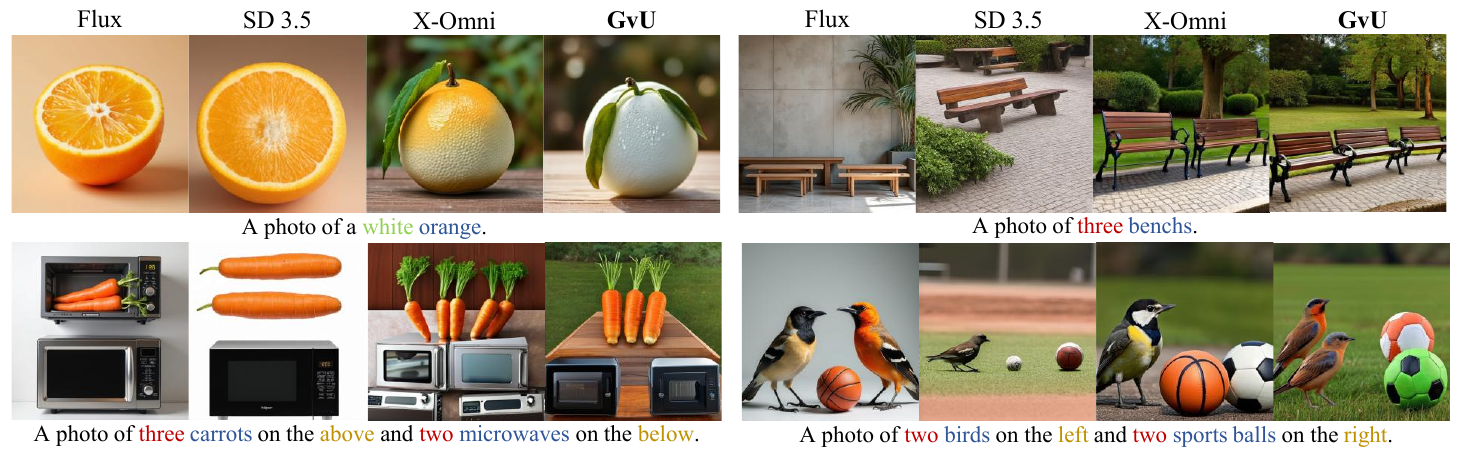}
    \vspace{-22pt}
    \caption{Qualitative comparisons between our GvU and other methods. Compared to other approaches, GvU generates images with better text-image alignment and more coherent spatial layout.}
    \label{fig:qualitative_figure}
\vspace{-13pt}
\end{figure*}

\subsection{Self-Supervised RL Optimization}
\label{sec_method_rl}

To bridge the gap between understanding and generation in UMMs, we design a self-supervised reinforcement learning (RL) framework with token-level intrinsic reward, optimized via Group Relative Policy Optimization (GRPO)~\cite{shao2024deepseekmath} algorithm. This framework provides fine-grained confidence estimates for text–image alignment, facilitating smoother optimization and more precise modeling of complex T2I instruction-following behaviors.

We implement a self-supervised GRPO training strategy, which eliminates the need to maintain a value function and external reward models, improving computational efficiency. Specifically, for each textual prompt $T\sim \mathcal{D}_T$, a policy model $\pi_\theta$ generates a group of $G$ trajectories $\{o_1,o_2,\cdots,o_G\}$. In T2I scenarios, each $o_i$ corresponds to the generation process of an image $I_i$. After generating images $\{I_1,I_2,\cdots, I_G\}$, GvU provides a group of model-intrinsic reward proposed in Sec.~\ref{sec_method_reward}:
\begin{equation}
    R_i(T,I_i)= P(T|I_i),\quad i=1,2,\cdots,G.
\end{equation}

The estimation of an advantage $A_i$ for each generated process is:
\begin{equation}
    A_i= \frac{R_i(T,I_i)- \text{mean}(\{R_i(T,I_i)\}_{i=1}^G)}{\text{std}(\{R_i(T,I_i)\}_{i=1}^G)}.
\end{equation}

The policy model $\theta$ is updated by maximizing the GRPO objective function:
\begin{equation}
\begin{aligned}
\mathcal{J}_{\text{GRPO}} & (\theta)
= \mathbb{E}_{T \sim \mathcal{D}_T,\, \{o_i\}_{i=1}^G \sim \pi_{\text{old}}(\cdot|T)} \\
& \Bigg[
    \frac{1}{G} \Big( \sum_{i=1}^G 
    \min \Big(
        r_i(\theta) A_i, \,
        \text{clip}\big(r_i(\theta), 1-\epsilon, 1+\epsilon\big) A_i
    \Big)
\\
& - \beta \,  
 \mathbb{D}_{\text{KL}}\big(\pi_{\theta}(\cdot|T) \, \| \, \pi_{\text{ref}}(\cdot|T)\big)
 \Big)
\Bigg], 
\end{aligned}
\end{equation}
where $r_i(\theta)= \frac{\pi_\theta(o_i|T)}{\pi_{\theta_{old}}(o_i|T)}$ is the probability ratio between the current policy model $\theta$ and old policy model $\theta_{old}$, $\epsilon$ and $\beta$ are hyper-parameters, and $\mathbb{D}_{\text{KL}}$ is the Kullback–Leibler divergence~\cite{shlens2014notes} to balance reward maximization and deviations from the reference model.

%% file: sec/5_experiments.tex
\section{Experiments}

In this section, we systematically evaluate our proposed intrinsic reward mechanism, GvU, from three complementary perspectives: \textit{performance}, \textit{learning dynamics}, and \textit{capability balance}. 
Specifically, we investigate:
1) the performance of GvU on visual generation tasks;
2) the cumulative effects that emerge progressively throughout the reinforcement learning process; 
3) the synergistic effect between generation enhancement and visual understanding, i.e., how improvements in generative performance influence fine-grained visual comprehension.

\begin{table*}[]
\caption{Evaluation of long-form T2I performance on DPG-Bench~\cite{hu2024dpg} benchmark. \textbf{Bold} indicates best, {\ul underline} indicates second best.}
\vspace{-9pt}
\centering
\begin{tabular}{@{}c|c|ccccc|c@{}}
\toprule
Model Type & Method & Global $\uparrow$ & Entity $\uparrow$ & Attribute $\uparrow$ & Relation $\uparrow$ & Other $\uparrow$ & Overall $\uparrow$ \\ \midrule
 & SD XL~\cite{podell2023sdxl} & 83.27 & 82.43 & 80.91 & 86.76 & 80.41 & \cellcolor[HTML]{EFEFEF}74.65 \\
 & DALL-E~\cite{ramesh2022hierarchical} & 90.97 & 89.61 & 88.39 & 90.58 & 89.83 & \cellcolor[HTML]{EFEFEF}83.50 \\
 & SD3-medium~\cite{esser2024scaling} & 87.90 & 91.01 & 88.83 & 80.70 & 88.68 & \cellcolor[HTML]{EFEFEF}84.08 \\
 & FLUX.1-dev~\cite{flux2024} & 82.1 & 89.5 & 88.7 & 91.1 & 89.4 & \cellcolor[HTML]{EFEFEF}84.0 \\
\multirow{-5}{*}{Only Generation} & OmniGen~\cite{xiao2025omnigen} & 87.90 & 88.97 & 88.47 & 87.95 & 93.56 & \cellcolor[HTML]{EFEFEF}81.16 \\ \midrule
 & Show-o~\cite{xie2024show} & 79.33 & 75.44 & 78.02 & 84.45 & 60.80 & \cellcolor[HTML]{EFEFEF}67.27 \\
 & Emu3~\cite{wang2024emu3} & 85.21 & 86.68 & 86.84 & 90.22 & 83.15 & \cellcolor[HTML]{EFEFEF}80.60 \\
 & Janus Pro~\cite{chen2025januspro} & 86.90 & 88.90 & 89.40 & 89.32 & 89.48 & \cellcolor[HTML]{EFEFEF}84.19 \\
 & UniWorld-V1~\cite{lin2025uniworld} & 83.64 & 88.39 & 88.44 & 89.27 & 87.22 & \cellcolor[HTML]{EFEFEF}81.38 \\
 & BLIP3-o 8B~\cite{chen2025blip3o} & - & - & - & - & - & \cellcolor[HTML]{EFEFEF}81.60 \\
 & OmniGen2~\cite{wu2025omnigen2} & 88.81 & 88.83 & 90.18 & 89.37 & 90.27 & \cellcolor[HTML]{EFEFEF}83.57 \\
 & BAGEL~\cite{deng2025bagel} & 88.94 & 90.37 & 91.29 & 90.82 & 88.67 & \cellcolor[HTML]{EFEFEF}{\ul 85.07} \\
 & XOmni~\cite{geng2025xomni} & 80.24 & 91.03 & 86.95 & 92.84 & 80.80 & \cellcolor[HTML]{EFEFEF}84.08 \\
\multirow{-9}{*}{\begin{tabular}[c]{@{}c@{}}Generation\&\\ Understanding\end{tabular}} & \cellcolor[HTML]{EFEFEF}GvU (Ours) & \cellcolor[HTML]{EFEFEF}85.28 & \cellcolor[HTML]{EFEFEF}91.38 & \cellcolor[HTML]{EFEFEF}90.25 & \cellcolor[HTML]{EFEFEF}93.88 & \cellcolor[HTML]{EFEFEF}85.20 & \cellcolor[HTML]{EFEFEF}\textbf{85.68} \\ \bottomrule
\end{tabular}
\label{table:dpg_bench}
\vspace{-12pt}
\end{table*}


\subsection{Experimental Setup}

\noindent \textbf{Model Architecture}. We employ the unified multimodal model (UMM) with an autoregressive (AR) and diffusion head architecture, where the AR module processes semantic tokens and the diffusion head reconstructs them into full-resolution images. Specifically, we employ the pretrained parameters of X-Omni~\cite{geng2025xomni}, a model trained using the SigLip-VQ tokenizer~\cite{zhai2023sigmoid}, Qwen2.5~\cite{qwen2025qwen25technicalreport}, and FLUX~\cite{flux2024}. We apply self-supervised reinforcement learning with GvU to the regular version of X-Omni. In Sec.~\ref{sec:exp_base_model}, we further train a weak base of X-Omni using GvU.



\noindent \textbf{Baselines}. We compare models trained by GvU with models on both visual generation and understanding. 
For generation, GvU is compared with specialized T2I models and UMMs.
For understanding, GvU is compared with dedicated understanding models and UMMs. All the compared models can be found in Table~\ref{table:geneval}, ~\ref{table:dpg_bench}, ~\ref{table:genevalplus}, and ~\ref{table:understanding}.



\begin{figure*}[!htbp]
    \centering
    \includegraphics[width=1.0\linewidth]{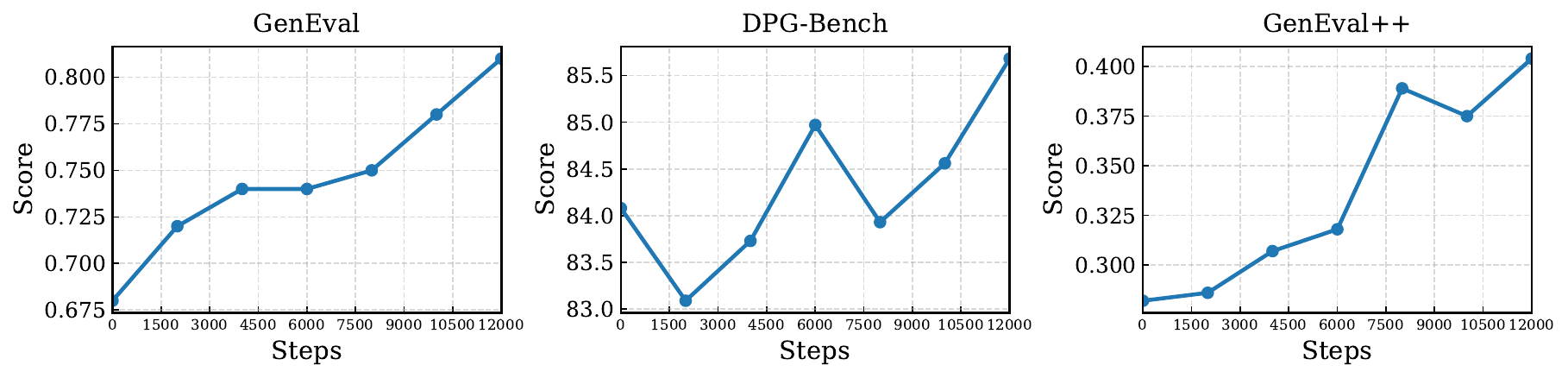}
    \vspace{-24pt}
    \caption{Evolution across multiple benchmarks during the self-supervised RL process. As the RL training steps increase, our model demonstrates steadily improved visual generation performance.}
    \label{fig:training_curves}
\vspace{-15pt}
\end{figure*}

\noindent \textbf{Datasets and Benchmarks}.
We use a text-only prompt dataset as the training set, comprising 50,000 samples that describe objects with positional relationships, quantitative relationships, and fine-grained details.
We evaluate our method on visual generation benchmarks, including GenEval~\cite{ghosh2023geneval}, DPG-Bench~\cite{hu2024dpg}, and GenEval++~\cite{ye2025echo}. 
GenEval evaluates compositional generation from an object-centric perspective, measuring object co-occurrence, position, count, and color using automated detectors aligned with human judgment. DPG-Bench focuses on dense, semantically rich prompts to test whether models can correctly generate multiple objects, attributes, and relationships. Geneval++ further scales GenEval to prompts with three or more objects and more complex spatial constraints, with image quality assessed by GPT-4.1.
For visual understanding, we evaluate on POPE~\cite{li2023evaluating}, GQA~\cite{hudson2019gqa}, MMB~\cite{liu2024mmbench}, SEED~\cite{li2024seedbench}, DocVQA~\cite{mathew2021docvqa}, and OCRB~\cite{liu2024ocrbench}, covering object recognition, relational reasoning, and document-based QA across diverse domains.
We further use MMT-Bench~\cite{ying2024mmtbench} and its fine-grained visual understanding sub-tasks to assess how enhancing generation affects fine-grained visual comprehension.

\noindent \textbf{Implementation Details}. We notice that open-source reinforcement learning frameworks for UMMs are currently scarce. To address this gap, we develop a GRPO-based RL training framework for UMMs, built upon TRL~\cite{vonwerra2022trl}. During training, the models are trained with LoRA~\cite{hu2022lora}. 
\subsection{Visual Generation Evaluation}

\noindent \textbf{Quantitative and Qualitative Evaluation}.
We compare our model with generation-only models and UMMs on the GenEval~\cite{ghosh2023geneval}, DPG-Bench~\cite{hu2024dpg}, and GenEval++~\cite{ye2025echo} benchmarks, which are reported in Table~\ref{table:geneval}, Table~\ref{table:dpg_bench}, and Table~\ref{table:genevalplus}, respectively.
On GenEval, our method reaches a score of 0.84. After GvU training, the model’s performance improves from 0.68 to 0.81 (without LLM rewriting), representing a 19.1\% relative increase, demonstrating superior performance.
On DPG-Bench, our method attains a competitive score of 85.68. It demonstrates particularly strong performance in Entity and Relation subcategories, indicating that the UMM’s understanding module effectively leverages spatial and relational comprehension to enhance the generative process.
On GenEval++, our method achieves a score of 0.404, showing a more pronounced performance with 43.3\% improvement after GvU training (0.282 to 0.404). Notably, it exhibits significant gains in mixed-category evaluations (pos/count, pos/size, color/pos), suggesting that the token-level fine-grained reward effectively captures intricate textual instruction signals and thereby facilitates the generation of more complex and compositionally consistent images.
Category-wise analysis reveals substantial gains in two-object generation and color accuracy, which may be attributed to the strong color perception capability of the understanding branch of the model.
Finally, we provide qualitative comparisons in Figure~\ref{fig:qualitative_figure}, which indicate that GvU generates images that are visually consistent with the input captions and exhibit coherent compositional organization among the depicted elements.

\begin{table*}[!htbp]
\caption{Evaluation of T2I performance on GenEval++~\cite{ye2025echo} benchmark. \textbf{Bold} indicates best, {\ul underline} indicates second best.}
\vspace{-9pt}
\centering
\resizebox{\textwidth}{!}{
\begin{tabular}{@{}c|c|ccccccc|c@{}}
\toprule
Model Type & Method & Color $\uparrow$& Count  $\uparrow$& Color/Count  $\uparrow$& Color/Pos  $\uparrow$& Pos/Count  $\uparrow$& Pos/Size  $\uparrow$& Multi-Count  $\uparrow$& Overall  $\uparrow$\\ \midrule
 & SD v2.1~\cite{rombach2022high} & 0.000 & 0.325 & 0.025 & 0.000 & 0.000 & 0.025 & 0.075 & \cellcolor[HTML]{EFEFEF}0.064 \\
 & SD XL~\cite{podell2023sdxl} & 0.050 & 0.375 & 0.000 & 0.000 & 0.000 & 0.000 & 0.000 & \cellcolor[HTML]{EFEFEF}0.061 \\
 & SD3-medium~\cite{esser2024scaling} & 0.550 & 0.500 & 0.125 & 0.350 & 0.175 & 0.150 & 0.225 & \cellcolor[HTML]{EFEFEF}0.296 \\
\multirow{-4}{*}{Only Generation} & FLUX.1-dev~\cite{flux2024} & 0.350 & 0.625 & 0.150 & 0.275 & 0.200 & 0.375 & 0.225 & \cellcolor[HTML]{EFEFEF}0.314 \\ \midrule
 & Janus-Pro~\cite{chen2025januspro} & 0.450 & 0.300 & 0.125 & 0.300 & 0.075 & 0.350 & 0.125 & \cellcolor[HTML]{EFEFEF}0.246 \\
 & BLIP3-o 8B~\cite{chen2025blip3o} & 0.250 & 0.250 & 0.125 & 0.600 & 0.125 & 0.575 & 0.225 & \cellcolor[HTML]{EFEFEF}0.307 \\
 & OmniGen2~\cite{wu2025omnigen2} & 0.550 & 0.425 & 0.200 & 0.275 & 0.125 & 0.250 & 0.450 & \cellcolor[HTML]{EFEFEF}0.325 \\
 & BAGEL~\cite{deng2025bagel} & 0.325 & 0.600 & 0.250 & 0.325 & 0.250 & 0.475 & 0.375 & \cellcolor[HTML]{EFEFEF}{\ul 0.371} \\
 & XOmni~\cite{geng2025xomni} & 0.225 & 0.500 & 0.025 & 0.325 & 0.150 & 0.475 & 0.275 & \cellcolor[HTML]{EFEFEF}0.282 \\
\multirow{-6}{*}{\begin{tabular}[c]{@{}c@{}}Generation\&\\ Understanding\end{tabular}} & \cellcolor[HTML]{EFEFEF}GvU (Ours) & \cellcolor[HTML]{EFEFEF}0.300 & \cellcolor[HTML]{EFEFEF}0.400 & \cellcolor[HTML]{EFEFEF}0.150 & \cellcolor[HTML]{EFEFEF}0.575 & \cellcolor[HTML]{EFEFEF}0.525 & \cellcolor[HTML]{EFEFEF}0.675 & \cellcolor[HTML]{EFEFEF}0.400 & \cellcolor[HTML]{EFEFEF}\textbf{0.404} \\ \bottomrule
\end{tabular}
}
\label{table:genevalplus}
\vspace{-12pt}
\end{table*}

\begin{figure*}[]
    \centering
    \includegraphics[width=1.0\linewidth]{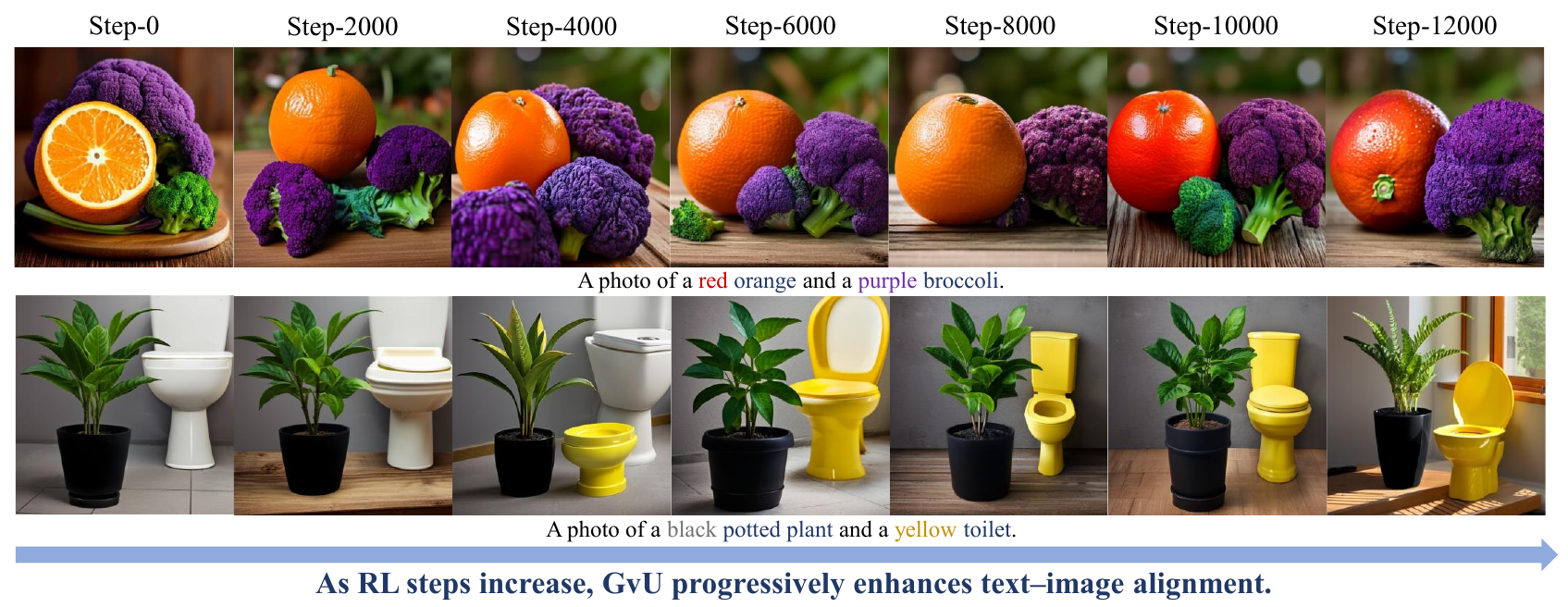}
    \vspace{-18pt}
    \caption{Illustration of GvU’s generated results across training steps. With ongoing self-supervised RL, our model effectively leverages intrinsic rewards to progressively enhance the text-image alignment, demonstrating continual improvement in T2I tasks.}
    \label{fig:step_figure}
\vspace{-10pt}
\end{figure*}

\noindent \textbf{Cumulative Effects of T2I Performance}.
One key question is whether the intrinsic reward enhances the generation performance of the UMM progressively or through a sudden leap. To explore this, we analyze the evolution of image generation quality throughout training, as illustrated in Figure~\ref{fig:training_curves}. 
By examining the relationship between training steps and performance across three benchmarks, we observe a steady and consistent improvement under the guidance of the intrinsic reward.
This indicates a cumulative effect of fine-grained intrinsic rewards, where token-level probability provides dense and informative feedback that drives continuous and robust performance gains in complex T2I tasks.
Qualitative results, illustrated in Figure~\ref{fig:step_figure}, further support this observation. As RL steps increases, GvU generates images that are progressively better aligned with the input text, exhibiting increasingly coherent and semantically consistent compositional layouts.

\vspace{-2pt}
\subsection{Visual Understanding Evaluation}
\vspace{-2pt}

We assess whether enhancing visual generation via intrinsic reward also influences visual understanding. As shown in Table~\ref{table:understanding}, our GvU achieves performance comparable with other models on six image understanding benchmarks, despite being trained without any image-understanding supervision. Strikingly, on MMT-Bench~\cite{ying2024mmtbench} (Table~\ref{tab:mmtbench}), which includes subtasks requiring fine-grained visual understanding—Visual Recognition (VR), Visual Illusion (VI), Hallucination (Ha), Visual Commonsense Reasoning (VCR), and Discipline Knowledge Reasoning (DKR)—we observe clear improvements. This indicates that enhancing generative capabilities can in turn facilitate fine-grained visual understanding, highlighting the potential of UMMs to realize the synergy between generation and understanding.




\subsection{Ablation on Base Models}\label{sec:exp_base_model}

\begin{figure}[]
    \centering
    \includegraphics[width=1\linewidth]{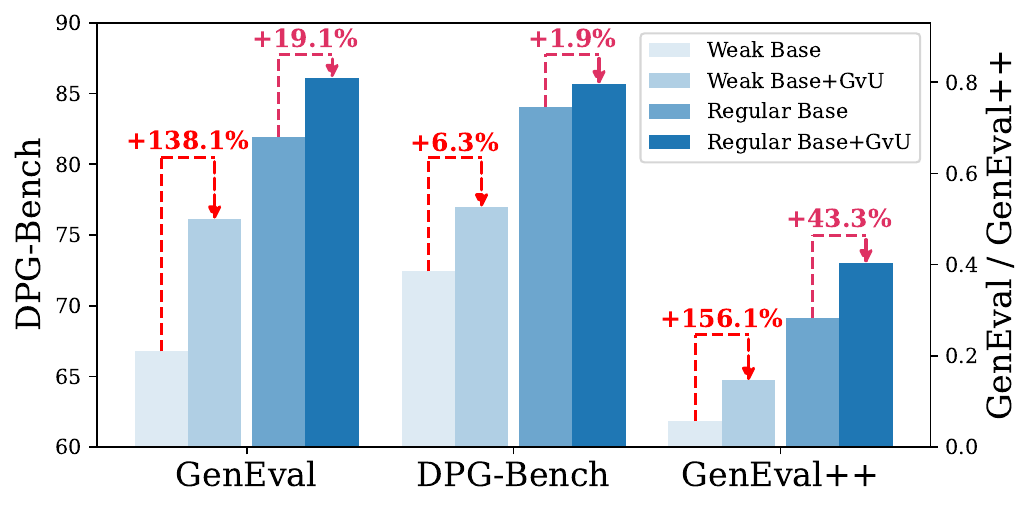}
    \vspace{-21pt}
    \caption{Comparison of GvU improvements on the weak and regular base models. GvU achieves greater gains on the weak base, suggesting that a larger gap between generation and understanding enhances its effectiveness in model training.}
    \label{fig:zhuzhuangtu}
\vspace{-12pt}
\end{figure}

Our intrinsic reward leverages the gap between the UMM’s understanding and generation capabilities, raising the question of how the gap’s magnitude affects the understanding branch’s guidance over generation.
To explore this, we further train the weak base of X-Omni~\cite{geng2025xomni} using GvU. Compared with the regular base, the weak base exhibits slightly stronger understanding ability (50.69 vs. 49.76 on MMT-Bench~\cite{ying2024mmtbench}) but weaker generation ability (0.21 vs. 0.68 on GenEval~\cite{ghosh2023geneval}), providing a suitable setup to examine the effect of gap size. As shown in Figure~\ref{fig:zhuzhuangtu}, GvU achieves a notably larger improvement on weak base with a wider gap compared with regular base (e.g., +138.1\% vs. +19.1\% on GenEval). Category-wise analyses further support it—for example, the multicount score in GenEval++ increases from 0 to 0.175, indicating that the intrinsic reward transfers generative knowledge previously absent in the weak base. 

\subsection{Ablation on Intrinsic Reward}
\begin{figure}[]
    \centering
    \includegraphics[width=1.0\linewidth]{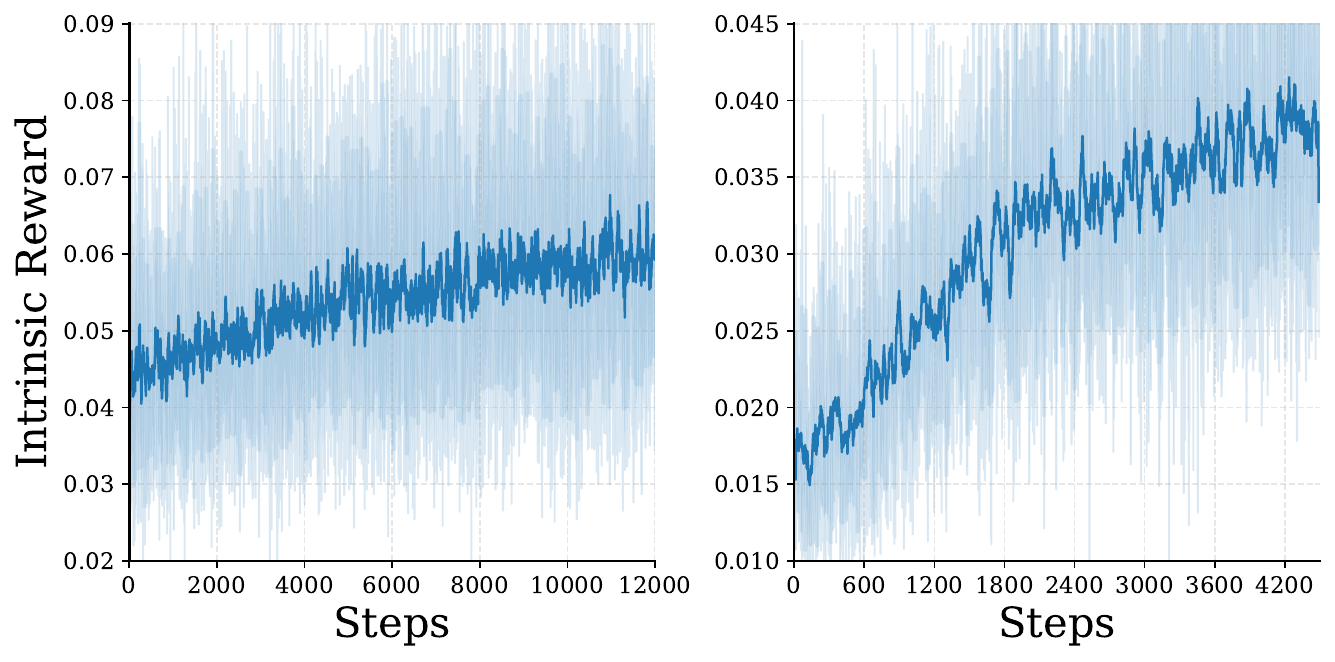}
    \vspace{-20pt}
    \caption{Reward training curves of GvU.  (Left) The left panel shows training starting from the regular base, which starts with higher initial rewards and supports more training steps.
    (Right) The right panel starts from the weak base, which begins lower but exhibits a larger relative increase in rewards.}
    \label{fig:reward_curve}
\vspace{-10pt}
\end{figure}

\noindent \textbf{Intrinsic Reward Training Curves}. Figure~\ref{fig:reward_curve} shows RL training curves of GvU.
Results demonstrate that the self-supervised RL framework consistently drives a smooth and stable increase in reward.
The regular base starts with a higher intrinsic reward (0.045) compared to the weak base (0.018), reflecting the stronger initial generation capability of the regular base.
This demonstrates that the intrinsic reward effectively captures generation quality and that the model’s internal probability distributions serve as a reliable proxy for image fidelity.
Moreover, the regular base can be trained continuously up to 12,000 steps, and training from the weak base exhibits a larger proportional improvement (0.018 $\rightarrow$ 0.038). This indicates that models with lower initial capabilities benefit more from intrinsic-reward-guided training, highlighting the framework’s ability to promote learning under limited initial generation performance.

\begin{table}[]
\caption{Evaluation on visual understanding benchmarks.}
\vspace{-9pt}
\centering
\resizebox{1\linewidth}{!}{
\begin{tabular}{@{}c|cccccc@{}}
\toprule
Method & POPE $\uparrow$ & GQA $\uparrow$ & MMB $\uparrow$ & SEED $\uparrow$ & DocVQA $\uparrow$ & OCRB $\uparrow$ \\ \midrule
LLaVA-NeXT~\cite{liu2024llavanext} & 86.1 & 64.5 & 66.9 & 68.6 & 74.1 & 529 \\
LLaVA-OneVision~\cite{li2024llavaonevision} & - & - & 80.8 & 75.4 & 87.5 & 622 \\
Janus-Pro~\cite{chen2025januspro} & 86.7 & 61.3 & 78.2 & 71.9 & - & 601 \\
Emu3~\cite{wang2024emu3} & 84.5 & 59.8 & 57.9 & 68.3 & 76.4 & 692 \\
X-Omni~\cite{geng2025xomni} & 86.1 & 61.9 & 73.9 & 74.5 & 89.1 & 712 \\
\textbf{GvU (Ours)} & 86.3 & 61.7 & 73.5 & 73.9 & 88.4 & 709 \\ \bottomrule
\end{tabular}
}
\label{table:understanding}
\vspace{-8pt}
\end{table}

\begin{figure}[]
    \centering
    \includegraphics[width=1.0\linewidth]{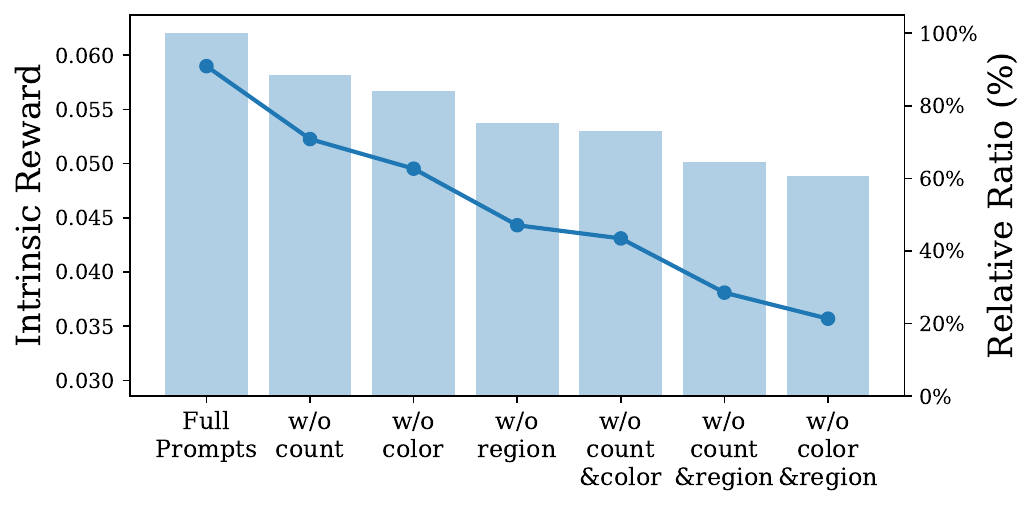}
    \vspace{-20pt}
    \caption{Sensitivity of intrinsic reward to fine-grained semantics. The line chart depicts the intrinsic reward, while the bar chart shows the reward relative to that of full prompts. Removing detail-specific terms causes a drop in intrinsic rewards, demonstrating that GvU effectively captures fine-grained semantic correspondences between text and images.}
    \label{fig:ablation_score_value}
\vspace{-14pt}
\end{figure}

\noindent \textbf{Intrinsic Reward Semantics Sensitivity}. To evaluate the sensitivity of the intrinsic reward to evaluate text–image alignment, we systematically remove count, color, and region terms from the evaluation prompts and compute the intrinsic reward between the modified prompts and the generated images. As shown in Figure~\ref{fig:ablation_score_value}, the reward consistently decreases once these detailed descriptors are removed, with the largest decline observed for region terms. Furthermore, removing more descriptors leads to larger score drops. These results indicate that the intrinsic reward effectively reflects fine-grained text–image alignment.

\vspace{-3pt}
\section{Limitation and Future Work}
\vspace{-3pt}
Achieving mutual enhancement between understanding and generation is a core goal of UMMs. We show that understanding can enhance generation, and improved generation can in turn strengthen fine-grained visual understanding. 
Nevertheless, the gains in understanding is still relatively modest, suggesting that further efforts are needed to more fully close this gap. We leave it for future work.

\begin{table}[]
\caption{Evaluation on overall score and fine-grained visual understanding subtasks of MMT-Bench~\cite{ying2024mmtbench}. }
\vspace{-9pt}
\centering
\resizebox{1\linewidth}{!}{
\begin{tabular}{@{}c|c|ccc|cc@{}}
\toprule
\multirow{2}{*}{Model} & MMT-Bench & \multicolumn{3}{c|}{Visual Details} & \multicolumn{2}{c}{Visual Reasoning} \\ \cmidrule(l){2-7} 
 & Overall & VR & VI & Ha & VCR & DKR \\ \midrule
Base & 49.76 & 51.21 & 45.57 & 66.25 & 70.0 & 38.46 \\
GvU & 49.92 ({\color[HTML]{1F77B4} +0.16}) & 52.58 ({\color[HTML]{1F77B4} +1.37}) & 50.63 ({\color[HTML]{1F77B4} +5.06}) & 68.75 ({\color[HTML]{1F77B4} +2.5}) & 75.0 ({\color[HTML]{1F77B4} +5.0}) & 42.31 ({\color[HTML]{1F77B4} +3.85}) \\ \bottomrule
\end{tabular}
}
\label{tab:mmtbench}
\vspace{-8pt}
\end{table}

%% file: sec/6_conclusion.tex
\section{Conclusion}

This paper proposes GvU, an intrinsic reward mechanism that exploits the understanding–generation gap in UMMs. Together with a self-supervised reinforcement learning strategy, GvU enables UMMs to gradually acquire complex T2I capability without additional annotations. The key idea is to use the model’s understanding branch to construct a self-consistent reward signal that quantifies the semantic discrepancy between understanding and generation, thus providing fine-grained guidance for improving generation quality. Experiments demonstrate that GvU consistently enhances T2I performance across multiple benchmarks. Furthermore, improving generation capability in turn strengthens the model’s fine-grained understanding. These results reveal a dynamic synergy between understanding and generation in UMMs, offering a foundation for developing truly unified multimodal systems.

\section{Acknowledgement}

This work was supported by the National Nature Science Foundation of China (6232221, 62525201, 62132001, 62432001) and Beijing Natural Science Foundation (L247006), the "Pioneer" and "Leading Goose" R\&D Program of Zhejiang Province(2024C01023).


